\documentclass[10pt,twocolumn,letterpaper]{article}
\usepackage[pagenumbers]{cvpr}
\usepackage{microtype}
\usepackage{mathtools}
\usepackage{multirow}
\newcommand{\refs}{\mathcal{R}}

\definecolor{cvprblue}{rgb}{0.21,0.49,0.74}
\usepackage[pagebackref,breaklinks,colorlinks,allcolors=cvprblue]{hyperref}
\newcommand{\papertitle}{Visual Jev Rewards: Reference-Bound Verification\texorpdfstring{\\}{ }for Multi-Subject Image Generation}
\newcommand{\paperauthors}{Baoteng Li, Wenzhuo Wu, Kongming Liang, Zhanyu Ma}
\hypersetup{pdftitle={\papertitle},pdfauthor={\paperauthors}}
\def\confName{CVPR}
\def\confYear{2026}
\title{\papertitle}

\author{
Baoteng Li$^{1,2}$ \quad
Wenzhuo Wu$^{1,2}$ \quad
Kongming Liang$^{1,2,\dagger}$ \quad
Zhanyu Ma$^{1,2}$ \\
School of Artificial Intelligence, Beijing University of Posts and Telecommunications$^{1}$ \\
Beijing Key Laboratory of Multimodal Data Intelligent Perception and Governance$^{2}$ \\
{\tt\small meltry.lbt@bupt.edu.cn}
}

\begin{document}
\maketitle
\begin{abstract}
Multi-subject image generation requires rewards that verify whether requested attributes, actions, and relations hold for the specified reference subjects. Subject presence alone does not establish that the correct subjects participate in a requested interaction. We present reference-bound Visual Jev rewards that turn these visual decisions into generator training signals. Each subject-related question receives a positive label only when the requested condition and the relevant reference identities hold jointly. We construct fixed questions offline, train a Qwen3.5-4B verifier with binary supervision, and directly read Yes probabilities from its language-model head. Their mean supplies a GRPO reward while retaining individual judgments for inspection. Using 200 MICo-150K training tasks and 30 updates, the framework raises a GPT-5.4 composite score from 41.78 to 52.50 on a manually selected 897-task MICo-Bench subset; direct 27B rewards yield 51.84. Each reward is tested in one GRPO run, and offline human evaluation does not establish a statistically significant advantage over direct scoring. The study provides an initial implementation and evaluation of Visual Jev as a reference-bound reward for multi-subject image generation.
\end{abstract}

\section{Introduction}
\label{sec:intro}
Multi-subject image generation combines reference images and a text prompt to place specified people or objects into a new scene. Recent advances in multi-image conditioning and composition data have improved this capability~\cite{uno,mico}. Yet an image may contain both reference subjects while showing the reference person holding a different dog while the reference dog stands nearby. Subject presence and action presence can both be satisfied while the requested interaction fails. A reward for this task must judge whether the specified subjects are the participants in the specified conditions.

\begin{figure*}[t]
\centering
\includegraphics[width=\textwidth]{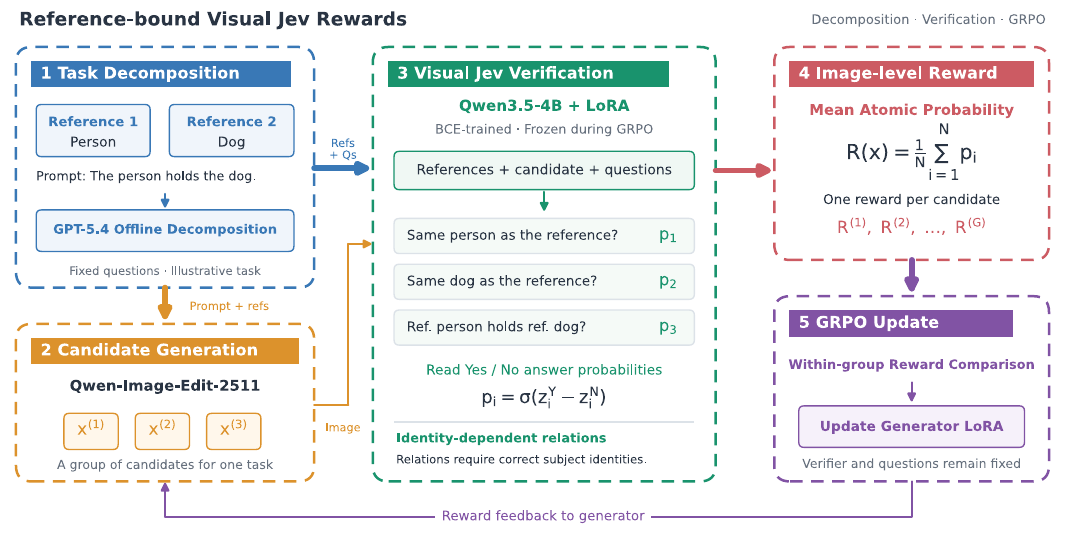}
\caption{\textbf{From Visual Jev decisions to generation rewards.} GPT-5.4 constructs fixed questions offline from the prompt and ordered references. A frozen, task-trained Visual Jev verifier reads Yes probabilities for each reference-bound condition; their arithmetic mean supplies the GRPO reward. Identity dependence is encoded in each relevant question and its positive-label semantics. Only the generator LoRA is updated during GRPO. The person-and-dog task is illustrative.}
\label{fig:method}
\end{figure*}

Reinforcement learning makes this reward design consequential: the reward determines which differences between generated candidates the model is trained to prefer. Flow-GRPO and PSR demonstrate the value of reward-based post-training for image generation and multi-subject personalization~\cite{flowgrpo,psr}. Question-based approaches, including TIFA, DSG, and AlphaGRPO, make requirements explicit through fine-grained verification~\cite{tifa,dsg,alphagrpo}. In multi-reference generation, a subject-related question must additionally identify which reference subject an attribute, action, or relation concerns. Scoring subject consistency and instruction following as separate dimensions can obscure whether the correct subjects actually participate in the requested interaction.

We study \textbf{reference-bound Visual Jev rewards for multi-subject image generation}. Jev introduces a decision interface that maps contextual questions to answer probabilities for programmatic use~\cite{jev,jevdocs}; Visual Jev applies candidate-answer probability readout to visual evidence~\cite{visualjev}. We bring this decision interface into the generation training loop. Our central design makes reference identity part of each relevant atomic judgment: a positive answer requires both the requested condition and the identities of its participants. We express this requirement in question wording and supervision labels, and train a local visual verifier to evaluate the resulting conditions.

The reward framework separates requirement construction, visual judgment, and reward aggregation. Questions are constructed offline from the prompt and ordered references, so every candidate for a task is judged against the same requirements. A task-trained Qwen3.5-4B verifier reads continuous Yes probabilities from its existing language-model output head. We average these decisions into an image-level reward for GRPO, retaining the individual scores for diagnosis. The verifier and question sets remain fixed as the generator is updated. Figure~\ref{fig:method} summarizes this training loop.

Our contributions are:
\begin{itemize}
\item \textbf{Reference-bound atomic verification.} We define subject-related conditions so that correct reference identity is necessary for a positive attribute, action, or relation judgment.
\item \textbf{A task-trained Visual Jev reward implementation.} We combine fixed task questions, binary visual supervision, and direct answer-probability readout into an inspectable reward for multi-reference candidates.
\item \textbf{An end-to-end generation study.} We integrate the frozen verifier into GRPO for Qwen-Image-Edit-2511 and evaluate generation, atomic verification, human preference agreement, and reward aggregation.
\end{itemize}

\section{Related Work}
\label{sec:related}
\subsection{Multi-Subject Generation and Composition}
UNO introduces multi-image conditioning through multi-subject data synthesis, progressive cross-modal alignment, and unified positional encoding, studying generalization from fewer to more subjects~\cite{uno}. MICo-150K supplies diverse multi-image composition data, a decomposition-and-recomposition subset, and MICo-Bench for evaluating subject preservation and complex composition~\cite{mico}. PSR applies reinforcement learning to multi-subject personalization using pairwise subject-consistency rewards together with general rewards~\cite{psr}. Our work studies the reward's joint judgment of identity and semantics: a requested action or relation must involve the subjects specified by the references.

\subsection{Atomic Visual Verification and Jev}
TIFA converts textual requirements into visual questions covering objects, counts, attributes, and relations~\cite{tifa}. DSG organizes such questions into a dependency graph to improve judgment consistency, including dependencies between object existence and attributes~\cite{dsg}. These studies establish question decomposition and dependent visual judgments as foundations for fine-grained evaluation. We build on them by specifying reference identity as a necessary condition within each relevant subject-related judgment.

TypeSafe AI's Jev is a System One decision model described as trained with Reinforcement Learning for Calibrated Decisions (RLCD)~\cite{jev}. Its Choice, Score, and Noul interfaces return structured decisions and probabilities, and its documentation recommends composing focused atomic judgments programmatically~\cite{jevdocs}. Visual Jev extends candidate-answer probability readout to visual tasks and studies answer supervision and shared visual-context execution~\cite{visualjev}. Recent RLJF work also explores Jev-style decision feedback as a GRPO reward for customer-support language models~\cite{rljf}. We study this reward interface for multi-subject image generation, grounding each relevant condition in reference images and training a visual verifier for the task. Our Qwen-based implementation follows the Jev decision interface; it uses BCE supervision rather than TypeSafe's RLCD training procedure.

\subsection{Verification Rewards for Image Synthesis}
Flow-GRPO applies online policy gradients to flow-matching generators, supporting reward-based optimization of compositional generation, text rendering, and preferences~\cite{flowgrpo}. EditScore shows the value of task-specific reward training for image editing~\cite{editscore}. Edit-R1 decomposes editing instructions into principles and scores them with a trained reasoning reward model~\cite{editr1}. AlphaGRPO's DVReward is especially close to our scoring interface: it constructs semantic and quality questions, reads normalized Yes/No probabilities, and aggregates them for reinforcement learning~\cite{alphagrpo}. We build on this question-based reward approach by defining identity-dependent conditions for multi-reference generation and training a dedicated visual decision model on them.

For multi-reference editing, EVR evaluates candidate-specific rationales and verifies them against visual evidence before forming rewards for DiffusionNFT~\cite{evr}. Our requirements are fixed from the task inputs before candidate generation, and a task-trained verifier directly predicts satisfaction scores for those same conditions across candidates. The contribution studied here is the use of reference-bound Visual Jev decisions in the generation training loop, with explicit positive-label semantics for the identities of participating subjects.

\section{Reference-Bound Visual Jev Rewards}
\label{sec:method}
Our framework turns reference-bound visual conditions into a reward for generation. It constructs fixed questions from the task inputs, learns to judge each complete condition with a visual decision model, and composes the resulting scores into a GRPO reward. The key design is the positive-label semantics of subject-related questions: the requested condition must hold for the correct reference subjects. Jev's question-driven decision interface and the visual answer-probability readout provide the implementation foundation~\cite{jev,jevdocs,visualjev}. Figure~\ref{fig:method} illustrates the training loop.

\subsection{Task Definition}
Given a text prompt $c$ and an ordered sequence of reference images $\refs=(r_1,\ldots,r_M)$, a generator $\pi_\theta$ produces an image $x$. The references specify visual identities, while the prompt specifies their attributes, actions, relations, and other scene requirements. We count reference images and subjects separately: a subject may have multiple reference images, and one reference may contain multiple subjects.

Let $u=(c,\refs)$ denote the input. We seek a reward $R_\phi(x,u)$ that provides a training signal for satisfying these requirements:
\begin{equation}
\max_\theta\; \mathbb{E}_{u\sim\mathcal{D}_{\mathrm{RL}},\,x\sim\pi_\theta(\cdot\mid u)}
\big[R_\phi(x,u)\big],
\end{equation}
where $\mathcal{D}_{\mathrm{RL}}$ contains the prompts and references used to train the generator, and $\phi$ denotes verifier parameters. We train the verifier first and fix $\phi$ during generator optimization.

\subsection{Atomic Questions Bound to References}
We use GPT-5.4 to construct questions offline from the prompt and ordered references:
\begin{equation}
Q(u)=\mathcal{G}(c,\refs)=\{q_i\}_{i=1}^{N_u},
\end{equation}
where $\mathcal{G}$ is the question-generation procedure and $N_u$ is the number of questions for the task. Decomposition uses only the task inputs, fixing the evaluation requirements before candidate generation. All candidates for a task share the same questions. Here, an atomic question checks one instruction-level requirement. It may include necessary identity conditions, so atomicity does not mean that the condition is logically indivisible.

Each question expresses a visually verifiable requirement and identifies its subjects through reference indices and necessary descriptions. For a prompt asking the reference person to hold the reference dog, separate questions can check the person's identity, the dog's identity, and the holding relation between those specified subjects. Interaction questions retain the participants' reference information so that both the interaction and its participants can be evaluated.

Question construction follows three principles. First, each question must be grounded in an input requirement, preserving requested identities, attributes, actions, and relations without turning incidental reference details into extra requirements. Second, each question must contain complete subject references and judgment conditions, allowing it to be answered individually from the references and candidate image. Third, the set should cover the main requirements while reducing semantic duplication, so that paraphrases do not count the same condition repeatedly.

We check format, reference citations, duplication, and conflicts with the source instruction, excluding or replacing tasks that cannot be interpreted unambiguously. Accepted questions are stored with the task and reused during reinforcement learning. Individual answerability requires self-contained wording; it does not imply logical independence among visual conditions.

\subsection{Identity-Dependent Judgment Criteria}
\textbf{Correct reference identity is part of the condition being judged.} Attributes, actions, and relations in multi-subject tasks concern specified subjects, so we include their identities in the positive-label semantics of the complete question. Let $\mathcal{S}_i$ be the set of reference subjects involved in $q_i$, let $I_s(x,\refs)\in\{0,1\}$ indicate whether the corresponding subject in the candidate matches reference subject $s$, and let $C_i(x)\in\{0,1\}$ indicate whether the specified attribute, action, or relation holds for the participants under consideration. For applicable subject-related questions, the positive-label semantics are
\begin{equation}
y_i=C_i(x)\;\land\!\bigwedge_{s\in\mathcal{S}_i} I_s(x,\refs).
\label{eq:identity}
\end{equation}
For example, ``Does the reference person hold the reference dog?'' is positive only if both identities and the holding relation are correct. If the candidate substitutes another dog, the requirement fails even when the holding action is present. The identity predicates refer to the participants in the queried interaction; the presence of a matching subject elsewhere in the image is insufficient. The same criterion applies to specified subjects' attributes and actions.

The identity constraint applies only to subjects involved in the question. An incorrect identity affects requirements involving that subject; other subjects and independent scene requirements are judged on their own evidence. Equation~\ref{eq:identity} defines the target label, not a guarantee that a learned verifier always implements it correctly. At inference, the verifier directly predicts a score for the complete question; we do not multiply separately predicted identity and relation probabilities or apply a separate identity gate. Records with insufficient evidence or annotation disagreement remain pending, and only confirmed binary labels are used for supervised training.

\subsection{Visual Jev Implementation and Training}
We implement the Visual Jev reward verifier using Qwen3.5-4B, LoRA, and binary cross-entropy (BCE), with candidate-answer probability readout from the language-model head~\cite{visualjev}. This is a local Qwen-based visual implementation inspired by Jev's decision interface. The verifier is trained with binary supervision, and GRPO is used to update the separate image generator. The verifier dataset contains candidate images, references, questions, and binary labels:
\begin{equation}
\mathcal{D}_{\mathrm{V}}=\{(x_n,\refs_n,q_n,y_n)\}_{n=1}^{K},\quad y_n\in\{0,1\}.
\end{equation}
A teacher model provides initial annotations, followed by consistency checks and human verification. Decomposition and answer annotation use different input scopes: decomposition defines requirements from the prompt and references, whereas answer annotation also observes the candidate image.

Given $(x,\refs,q_i)$, the verifier reads the Yes and No logits at the answer position, denoted by $z_i^Y$ and $z_i^N$. Normalizing over the two answers gives
\begin{equation}
p_i=\frac{\exp(z_i^Y)}{\exp(z_i^Y)+\exp(z_i^N)}
=\sigma(z_i^Y-z_i^N),
\label{eq:prob}
\end{equation}
where $\sigma$ is the sigmoid function and temperature is set to one. The value $p_i$ represents relative support for Yes in the binary decision. Direct logit readout supplies this score without generating a free-form rationale. We retain it for reward computation; its normalization alone does not establish probability calibration.

Writing $p_n=p_\phi(x_n,\refs_n,q_n)$, the verifier loss is
\begin{equation}
\begin{aligned}
\mathcal{L}_{\mathrm{V}}(\phi)=-\frac{1}{K}\sum_{n=1}^{K}\Big[&y_n\log p_n\\
&+(1-y_n)\log(1-p_n)\Big].
\end{aligned}
\end{equation}
Training updates LoRA parameters and retains the base model's language-model output head for answer readout. We then freeze the verifier for offline scoring and reinforcement learning. BCE trains the visual verifier; GRPO subsequently optimizes the image generator.

\subsection{Image-Level Reward Aggregation}
For candidate $x$ of task $u$, the verifier produces atomic scores $\mathbf{p}(x,u)=(p_1,\ldots,p_{N_u})$. We use their arithmetic mean as the image-level reward:
\begin{equation}
R_\phi(x,u)=\frac{1}{N_u}\sum_{i=1}^{N_u}p_\phi(x,\refs,q_i).
\label{eq:reward}
\end{equation}
This reward lies in $[0,1]$ and weights questions equally. Normalizing by question count prevents the reward magnitude from increasing directly with the number of questions. It measures the verifier's average support for the task requirements, so coverage and deduplication affect the resulting reward. Identity dependence enters through the judgment of each complete question, after which the scores are averaged. A subject's identity can recur in its identity question and in its associated attribute, action, or relation questions, so equal question weights need not imply equal subject weights.

The atomic mean retains evidence for diagnosis. Differences in individual $p_i$ values reveal which identity, attribute, or interaction conditions contribute to differences among candidate rewards. The mean does not guarantee simultaneous satisfaction of all requirements: high scores on other conditions can partly compensate for failure on a critical one.

\subsection{GRPO with Atomic Rewards}
We initialize the generator from the official Qwen-Image-Edit-2511 weights and add LoRA parameters for reinforcement learning. For task $u$, the sampling policy $\pi_{\theta_{\mathrm{old}}}$ generates a group of $G$ candidates:
\begin{equation}
\begin{aligned}
x^{(g)}&\sim\pi_{\theta_{\mathrm{old}}}(\cdot\mid u),\\
r^{(g)}&=R_\phi(x^{(g)},u),\qquad g=1,\ldots,G.
\end{aligned}
\end{equation}
Every candidate uses the same question set and scoring rule. The training procedure computes relative advantages from within-group reward differences and performs GRPO updates using image-generation trajectories, converting atomic satisfaction judgments into a generator training signal.

The questions and verifier parameters remain fixed, while the generator LoRA is updated. GPT-5.4 decomposition is completed before training, and the local Visual Jev verifier evaluates newly generated candidates during training. The verifier supplies one scalar reward per image, from which group-relative advantages are computed.

\section{Data and Experimental Setup}
\label{sec:setup}
\subsection{Datasets}
The atomic verification dataset contains 3,972 binary examples: 2,477 for training, 360 for development, 199 for calibration, and 936 for testing. It covers original candidates, candidate comparisons, and reference-replacement examples. All examples follow the same reference-bound question and identity-dependent labeling criteria, with GPT-5.4 annotation and human verification.

The GRPO data are selected from the MICo-150K training set and comprise 200 tasks, 549 reference images, and 1,634 atomic questions. GPT-5.4 constructs the questions offline from prompts and ordered references, and the questions remain fixed during training.

\subsection{Models and Training}
The Visual Jev verifier uses Qwen3.5-4B with LoRA and BCE. All offline reward and GRPO experiments use the checkpoint at update 136, selected by development-set balanced accuracy; its test balanced accuracy is 95.24\%. The verifier-objective comparison separately reports all trained methods at the fixed endpoint of update 544. Answer readout uses temperature one, and the binary classification threshold is 0.5. The image-level reward is the arithmetic mean of atomic Yes probabilities.

GRPO starts from the official Qwen-Image-Edit-2511 model, with LoRA rank 32, learning rate $5\times10^{-7}$, 12 candidates per group, $768\times768$ training resolution, 20 sampling steps during training, and CFG 4. Training completes 30 updates. The direct-reward control uses Qwen3.5-27B: instruction following and subject consistency are each scored from 0 to 5, and their sum is divided by 10. It uses the same training data and the settings above.

\subsection{Evaluation}
Atomic verification is measured using accuracy, balanced accuracy, negative-class recall, and the Brier score. Offline scoring compares Visual Jev with direct Qwen3.5-4B and Qwen3.5-27B scoring on the same fixed candidates and blinded human rankings. The development split has eight valid tasks; the validation split has 21 valid tasks and 126 strict preference pairs. We report strict preference agreement, Top-1 credit, and the risk of a severe identity error in the top-ranked candidate. The direct 27B scores shown in the offline case figure use a 0--100 scale and are divided by 100 for display. They belong to the offline scoring protocol; the GRPO control instead uses the two component scores described above.

For this initial downstream evaluation, human screening selects a subset of 897 higher-quality tasks from MICo-Bench, covering Object-Centric, Human-Centric, human--object interaction (HOI), and decomposition and recomposition (De\&Re). All three generator variants are evaluated on this common selected subset. Results therefore describe the screened collection rather than the full benchmark.

We use a GPT-5.4 composite score for this study, distinct from the official MICo-Bench evaluation protocol~\cite{mico}. GPT-5.4 scores naturalness, freedom from artifacts, instruction following, and subject similarity from 0 to 10. Perceptual quality and semantic consistency are
\begin{equation}
\begin{aligned}
\mathrm{PQ}&=\sqrt{s_{\mathrm{natural}}\cdot s_{\mathrm{artifact}}},\\
\mathrm{SC}&=\sqrt{s_{\mathrm{instruction}}\cdot s_{\mathrm{subject}}}.
\end{aligned}
\end{equation}
The composite score for each example is
\begin{equation}
S=\mathrm{PQ}\times\mathrm{SC},
\end{equation}
which ranges from 0 to 100. The overall score is the mean across the 897 selected tasks. The official Qwen-Image-Edit-2511 checkpoint and both GRPO models are evaluated using this same formula; the reported baseline score is our evaluation of the official weights.

\section{Experimental Results}
\label{sec:results}
We first test whether Visual Jev decisions provide useful rewards in the generation training loop. We then examine atomic verification, agreement with human preferences, and reward aggregation to characterize the behavior of the resulting reward.

\subsection{GRPO with Visual Jev Rewards}
Table~\ref{tab:grpo} reports generation performance on the manually selected 897-task MICo-Bench subset using our GPT-5.4 composite score. After 30 GRPO updates, the Visual Jev reward model yields a composite score of 52.50, improving over the official baseline of 41.78 by 10.72 points. Training with direct 27B rewards reaches 51.84, also above baseline. The two GRPO methods use the same training data, initialization, and main training hyperparameters. The Visual Jev method scores approximately 0.67 points higher in this evaluation, computed from unrounded means.

\begin{table}[t]
\centering
\caption{\textbf{Generation results on the screened MICo-Bench subset.} Baseline: official Qwen-Image-Edit-2511. Both GRPO models complete 30 updates. GPT-5.4 evaluates PQ$\times$SC on a 0--100 scale. Overall scores are averages over the same 897 manually selected tasks. One run is reported per method. Higher is better.}
\label{tab:grpo}
\small
\setlength{\tabcolsep}{3.2pt}
\begin{tabular}{lrrrrr}
\toprule
Model & Object & Human & HOI & De\&Re & Overall\\
\midrule
Baseline & 41.19 & 51.05 & 37.81 & 40.70 & 41.78\\
GRPO (27B) & 49.82 & 60.90 & 49.71 & \textbf{49.75} & 51.84\\
GRPO (Visual Jev) & \textbf{50.26} & \textbf{61.24} & \textbf{51.46} & 49.66 & \textbf{52.50}\\
\bottomrule
\end{tabular}
\end{table}

\begin{figure*}[t]
\centering
\includegraphics[width=\textwidth]{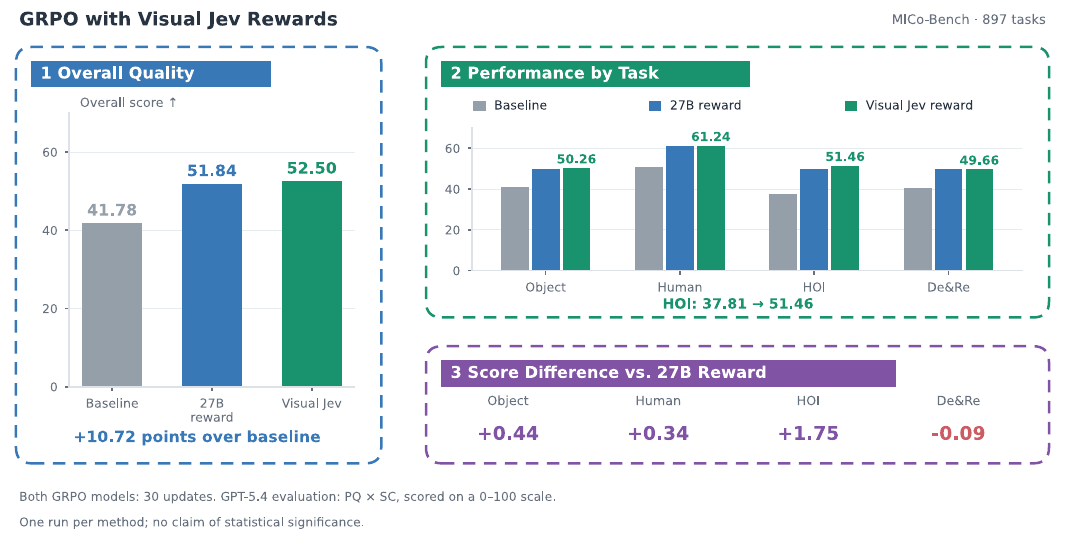}
\caption{\textbf{Overall and category-level GRPO results.} On the screened 897-task MICo-Bench subset, Visual Jev rewards improve our GPT-5.4 composite score by 10.72 points over the official Qwen-Image-Edit-2511 checkpoint. Relative to direct 27B rewards, the largest category gain is on HOI (+1.75); De\&Re is slightly lower ($-0.09$). Results are observations from one run per method, without a claim of statistical significance.}
\label{fig:grpo}
\end{figure*}

The Visual Jev-trained generator exceeds baseline in all four categories (Fig.~\ref{fig:grpo}). HOI increases from 37.81 to 51.46, a gain of 13.65 points. Compared with direct 27B rewards, Visual Jev gains 0.44 on Object, 0.34 on Human, and 1.75 on HOI, while decreasing by 0.09 on De\&Re. The larger HOI improvement is consistent with our focus on subjects and interactions, but the experiments do not isolate the identity-dependent rule as its cause.

These results indicate that a task-trained 4B verifier can supply useful GRPO rewards for multi-subject generation. The comparison evaluates complete trained systems and does not separate the contributions of verifier fine-tuning, question decomposition, and aggregation. Repeated training and component ablations are needed to determine whether the small observed difference from direct scoring is stable and which design choices account for it.

\begin{figure*}[t]
\centering
\includegraphics[width=\textwidth]{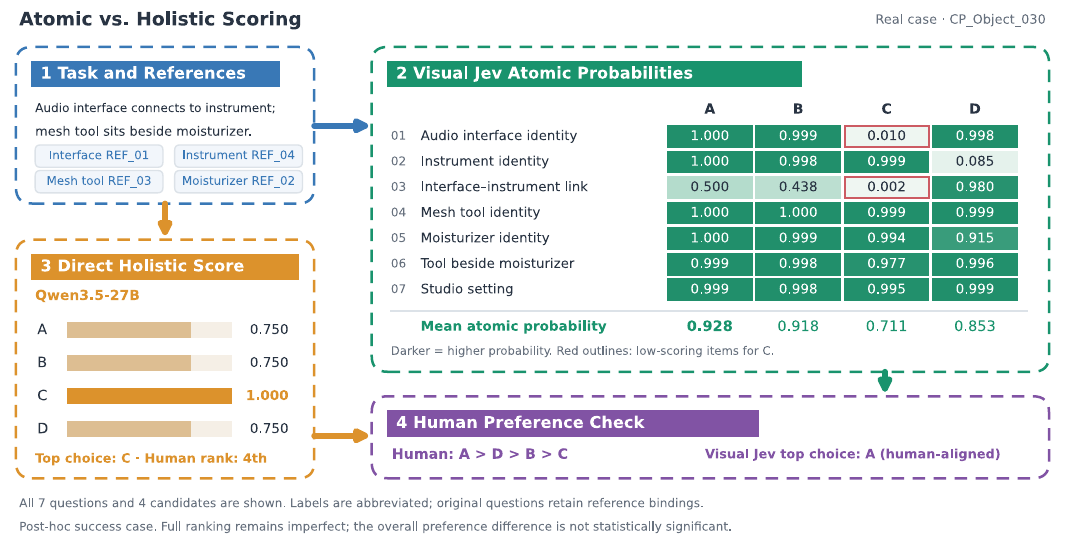}
\caption{\textbf{Atomic probabilities reveal the basis of a scoring disagreement.} All seven questions and four candidates from CP\_Object\_030 are retained; the displayed English labels abbreviate the reference-bound questions. Direct 27B scores are divided by 100 for display, without implying calibration equivalence to atomic probabilities. The direct scorer prefers C, which receives low Visual Jev probabilities for interface identity and the interface--instrument connection and is ranked last by the human. Visual Jev selects A, matching the human's first choice, although its full ranking remains imperfect. This post-hoc success case illustrates scoring evidence rather than overall performance.}
\label{fig:case}
\end{figure*}

\subsection{Atomic Verification}
Table~\ref{tab:atomic} compares verifier training objectives on 936 test examples and separately identifies the checkpoint used for rewards. The fixed-budget comparison uses update 544; all offline reward and GRPO results use the development-selected update-136 checkpoint. The untuned Qwen3.5-4B model uses the same binary probability readout, achieving 82.05\% accuracy and 84.14\% balanced accuracy. All three task-trained methods improve substantially. BCE reaches 96.15\% accuracy, a gain of 14.10 percentage points over the base model, and 96.04\% balanced accuracy, a gain of 11.90 points. Its Brier score decreases from 0.1332 to 0.0356. Supervised training on reference-bound questions therefore improves both condition judgments and probability predictions.

\begin{table}[t]
\centering
\caption{\textbf{Atomic verification under a fixed training budget.} The upper block compares trained methods at update 544 on 936 test examples. The lower row reports the development-selected reward checkpoint at update 136; dashes mark metrics not reported for that checkpoint. Acc.: accuracy; BA: balanced accuracy; Neg. R: negative-class recall. The first three metrics are percentages.}
\label{tab:atomic}
\small
\setlength{\tabcolsep}{4pt}
\begin{tabular}{lrrrr}
\toprule
Method & Acc.$\uparrow$ & BA$\uparrow$ & Neg. R$\uparrow$ & Brier$\downarrow$\\
\midrule
Untuned Qwen3.5-4B & 82.05 & 84.14 & 94.67 & 0.1332\\
Answer-token SFT & 96.05 & \textbf{96.04} & \textbf{96.00} & \textbf{0.0356}\\
BCE & \textbf{96.15} & \textbf{96.04} & 95.47 & \textbf{0.0356}\\
BCE + Brier & 95.73 & 95.60 & 94.93 & 0.0391\\
\midrule
BCE (reward, 136) & -- & 95.24 & -- & --\\
\bottomrule
\end{tabular}
\end{table}

SFT and BCE have the same balanced accuracy. SFT has higher negative-class recall, while BCE has slightly higher overall accuracy. Adding a Brier term does not improve the reported decision metrics or the Brier score in this setting. Brier score measures overall probability-prediction quality; a lower value alone does not isolate improved calibration. Subsequent reward experiments use the development-selected BCE checkpoint at update 136, whose test balanced accuracy is 95.24\%. The fixed-endpoint table compares objectives, whereas the deployed checkpoint reflects model selection.

\subsection{Human Preference Agreement}
Table~\ref{tab:preference} reports human evaluation on fixed candidates. Visual Jev achieves 63.49\% strict preference agreement, exceeding direct 27B scoring at 60.32\% and direct 4B scoring at 59.13\% by 3.17 and 4.37 percentage points, respectively. Its Top-1 credit is also higher, but severe identity risk does not improve correspondingly.

\begin{table}[t]
\centering
\caption{\textbf{Offline human preference comparison.} All methods use the same 21 valid tasks and 126 strict preference pairs. Direct 4B scoring is an additional control. All values are percentages; identity risk concerns the top-ranked candidate.}
\label{tab:preference}
\small
\setlength{\tabcolsep}{5pt}
\begin{tabular}{lrrr}
\toprule
Scoring method & Agreement$\uparrow$ & Top-1$\uparrow$ & Risk$\downarrow$\\
\midrule
Direct Qwen3.5-4B & 59.13 & 32.54 & 46.03\\
Direct Qwen3.5-27B & 60.32 & 39.29 & \textbf{45.24}\\
Visual Jev atomic mean & \textbf{63.49} & \textbf{42.86} & 47.62\\
\bottomrule
\end{tabular}
\end{table}

Paired bootstrap over parent tasks yields a 95\% confidence interval of $[-8.33,16.67]$ percentage points for the agreement difference against 27B and $[-2.38,11.51]$ against 4B. Both cross zero, so the current evidence shows a positive trend in mean agreement. After excluding tasks where every candidate is unacceptable, Top-1 credit is 33.33\% for Visual Jev and 37.50\% for 27B, indicating that the overall Top-1 advantage depends on the candidate quality distribution. High atomic accuracy does not by itself ensure that the aggregated reward reliably selects the best candidate with correct identities.

Figure~\ref{fig:case} provides a concrete example of the information retained by atomic scoring. The per-question probabilities expose identity and relation failures that can be obscured by a holistic score, while retaining the verifier's remaining errors for inspection.

\subsection{Reward Aggregation}
Table~\ref{tab:aggregation} compares three aggregation rules. On development data, the atomic mean and category mean both achieve 68.75\% strict preference agreement. The prespecified tie-breaking order selects the atomic mean. On validation data, the category mean and the mixture with the worst category both achieve 65.08\%, compared with 63.49\% for the atomic mean. We report the former two as secondary results; GRPO uses the atomic mean frozen during development.

\begin{table}[t]
\centering
\caption{\textbf{Offline aggregation comparison.} Development: eight valid tasks and 48 strict preference pairs. Validation: 21 valid tasks and 126 pairs. Values are preference agreement (\%). The mixture uses 0.75 times the category mean plus 0.25 times the worst-category mean.}
\label{tab:aggregation}
\small
\setlength{\tabcolsep}{8pt}
\begin{tabular}{lrr}
\toprule
Aggregation & Dev.$\uparrow$ & Val.$\uparrow$\\
\midrule
Atomic mean (selected) & \textbf{68.75} & 63.49\\
Category mean & \textbf{68.75} & \textbf{65.08}\\
Category/worst-category mixture & 66.67 & \textbf{65.08}\\
\bottomrule
\end{tabular}
\end{table}

These results suggest that the distribution of question categories affects image ranking. The atomic mean weights each question equally, so categories with more questions contribute more. Category averaging can adjust this effect. The current sample is too small to establish a consistently optimal aggregation rule, and the main method retains the simple, frozen atomic mean.

\section{Conclusion}
We formulate and implement reference-bound Visual Jev rewards for multi-subject image generation. Its central design binds subject-related visual conditions to the reference identities, making the correct participants part of each positive judgment. Fixed task questions, a task-trained visual verifier, and direct answer-probability readout turn these conditions into an image-level reward while retaining the individual decisions for inspection.

An initial GRPO study with 200 training tasks and 30 updates improves the GPT-5.4 composite score from 41.78 to 52.50 on the manually selected 897-task MICo-Bench subset. Atomic verification and offline human comparisons further characterize the reward's behavior. These results establish an initial working path from reference-bound visual decisions to generative model post-training. Larger independent evaluations and controlled ablations are needed to determine when this design offers a stable advantage and how much of the observed improvement comes from identity binding.

\section{Limitations}
\label{sec:limitations}
This study is an initial evaluation of the reward framework. The GRPO comparison uses one run per method and 30 updates, and the offline human validation contains 21 valid tasks. Preference-agreement confidence intervals cross zero, and the atomic reward does not reduce severe identity risk in the current offline comparison. The complete-system comparison does not isolate verifier training, question decomposition, identity-dependent labels, and aggregation.

The arithmetic mean permits strong scores on some requirements to compensate for a failed critical condition. The framework also depends on question coverage and the quality of binary annotations; generalization to broader subjects, tasks, and reference configurations requires further evaluation. GPT-5.4 participates in question construction, annotation, and downstream evaluation, so shared model preferences may influence the results. Independent human evaluation of the trained generators remains necessary. The manually screened 897-task subset favors higher-quality benchmark examples, which limits conclusions about the full benchmark. Downstream scores use the GPT-5.4 composite defined in Sec.~\ref{sec:setup} and are not presented as the benchmark's official evaluation protocol.

{\small
\bibliographystyle{ieeenat_fullname}
\bibliography{main}
}
\end{document}